\documentclass[preprint,12pt,authoryear]{elsarticle}
\usepackage{amsmath,amssymb}
\usepackage{booktabs,array}
\usepackage{graphicx}
\usepackage{xcolor}
\usepackage{tikz}
\usepackage{algorithm,algpseudocode}
\usetikzlibrary{positioning,arrows.meta}
\usepackage[hidelinks]{hyperref}
\journal{Data Science and Management}
\newif\ifanon \anonfalse

\begin{document}
\begin{frontmatter}
\title{Beyond fabrication: Misattributed numbers in papers written by AI research agents}

\ifanon
\else
\author[aau]{Mosab Alfaqeeh\corref{cor1}}
\ead{m.alfaqeeh@ammanu.edu.jo}
\author[aau]{Omar Mohammed Horani}
\ead{o.horani@ammanu.edu.jo}
\author[aau]{Mou'th Hourani}
\ead{mhourani@ammanu.edu.jo}
\author[bing]{Hashem Alfaqeeh}
\ead{halfageh@binghamton.edu}
\cortext[cor1]{Corresponding author.}
\affiliation[aau]{organization={Al-Ahliyya Amman University},
            city={Amman},
            country={Jordan}}
\affiliation[bing]{organization={Binghamton University, State University of New York},
            city={Binghamton},
            state={NY},
            country={USA}}
\fi

\begin{abstract}
Verifying numerical claims in research written by AI agents is an important but challenging problem. Agents can now run experiments and write complete papers, while organizations increasingly rely on their reported results. Recent benchmarks show that many agent-written papers fail integrity checks, but such verdicts, often produced by a single large language model (LLM) judge, identify failure without necessarily explaining its nature. Existing approaches either check a paper for internal consistency or test whether a reported value appears somewhere in an agent's output. Neither establishes whether the value comes from the experiment it is claimed to describe.

We re-analyze all 540 released judgments from a recent public benchmark of AI research agents, covering nine agents and fifteen open research problems, at the level of individual rubric criteria. We then manually label 256 numerical claims from 18 submissions, distinguishing claims with no supporting source in the available artifacts from values drawn from the wrong experiment, metric, or run. Failures concentrate on alignment between the paper and the underlying work rather than overt cheating. Among unsupported numerical claims in the labeled sample, approximately four in five correspond to values in the agent's files but lack support from the appropriate source, indicating that misattribution is a major failure mode in this benchmark.

To address this problem, we introduce a deterministic claim-to-evidence checker that scores whether the surrounding context identifies the appropriate result and calibrates each claim against nearby numerical decoys. With its operating rules fixed before any claims were labeled, the checker certifies claims with over 90\% precision against blinded human annotations (78 of 85 certified claims traceable, including 27 of 29 on held-out submissions; Cohen's $\kappa = 0.89$), while verifying only a subset of claims. An LLM verifier given the same evidence recovers more traceable claims but accepts about three times as many misattributed ones, so the two are best used together. Most claims it cannot verify are nevertheless judged traceable by human annotators, demonstrating why failure to verify should not be treated as evidence of falsehood. Applied to 99 submissions, the checker reveals substantial differences in numerical traceability across agents. On a sample of 30 submissions, a second judge configuration, with a different model and execution harness, orders the agents' vendor groups as the official judge does but passes twice as many submissions, showing that absolute integrity outcomes depend on the judge configuration. These findings motivate provenance-first auditing of AI-produced research and analytical reports.
\end{abstract}

\end{frontmatter}

\section{Introduction}\label{sec:intro}

Large language model (LLM) agents have moved from answering questions to doing extended analytical work: they write and run code, execute experiments, and report the results in documents that look like finished research \citep{lu2024aiscientist,chen2025mlrbench,starace2025paperbench}. For organizations, this changes where the risk sits. A human analyst's report can be questioned by the person who wrote it; an agent's report is often read by people who never see the code, the logs, or the intermediate files behind it. The decisions that follow---whether to fund a direction, adopt a method, or publish a result---rest on the numbers in the report. Whether those numbers can be traced to the work that supposedly produced them is therefore a data-quality and data-provenance question \citep{wang1996dataquality,simmhan2005provenance}, not only a question about model capability.

Recent benchmarks make the scale of the problem visible. On \emph{Into the Unknown} \citep{kim2026unknown}, nine agents each attempted fifteen open research problems drawn from recent papers, and an LLM judge audited every submission for integrity before assessing its scientific merit. Only 21 of the 135 submissions passed the integrity gate. Commentary on such results tends to describe the failures as fabrication: agents inventing data or results that were never produced. That description matters, because it implies a particular remedy---detecting invented numbers---and a particular level of distrust.

Two features of the evidence make that description premature. First, the verdicts come from a single LLM judge from one vendor, and LLM judges are known to be sensitive to presentation and, in some settings, to favor outputs resembling their own \citep{zheng2023judge,panickssery2024selfpref}. Second, an integrity verdict is an aggregate: it records \emph{that} a submission failed, not \emph{how}. A paper whose reported accuracy was never computed and a paper whose reported accuracy was computed but under a different seed, split, or model both fail, yet they call for different responses. The first is fabrication; the second is misreporting---the number exists, but the report attaches it to the wrong experiment.

This paper asks which of the two dominates, and what it takes to tell them apart at scale. We study four questions:
\begin{enumerate}
  \item[\textbf{RQ1}] What kinds of integrity failures does the official judge record, and do they point to fabrication or to misalignment between write-up and work?
  \item[\textbf{RQ2}] When a number in an agent's paper is not supported, is it absent from the agent's files, or present but drawn from the wrong source?
  \item[\textbf{RQ3}] Can numerical claims in AI-generated research papers be traced automatically to their supporting execution artifacts with useful precision, and without treating claims that cannot be verified as false?
  \item[\textbf{RQ4}] How robust are the benchmark's integrity conclusions to the choice of judge?
\end{enumerate}

To answer them, we combine four sources of evidence on the publicly released benchmark data. We re-analyze all 540 released judgments at the level of individual rubric criteria. We collect human labels for 256 numerical claims drawn from 18 submissions, using a development set and a separate, untouched confirmation set labeled independently by two annotators. We build a deterministic claim-to-evidence checker that matches each number in a paper against the agent's own result files and, crucially, tests whether the surrounding context identifies the right source rather than merely the right value. Finally, we re-grade a stratified sample of 30 submissions with a second judge configuration: a model from a different vendor, run in the benchmark's own judge runner with the official prompt, rubric, and grading code.

The evidence points consistently in one direction. The criteria that fail are those checking whether the write-up matches the work (code--paper alignment, number traceability), whereas criteria that would catch overt cheating (hard-coded values, external code, test-set leakage) almost never fire. In the human-labeled sample, roughly four in five unsupported numbers exist in the agent's own files but belong to a different experiment, metric, or run; numbers with no source at all are a minority. In this benchmark, in other words, misattribution of real numbers is a major failure mode, and outright absence of a source is the minority case.

This finding has a methodological consequence. If misreported numbers exist in the files, then checking whether a reported value appears anywhere in an agent's output---the obvious automated test---cannot detect them: in our data it accepts 87--95\% of misattributed numbers. Verification has to establish provenance, not just presence. Our checker does this by scoring how well the surrounding context identifies a candidate source and calibrating each claim against nearby decoy values. Its strict operating point was fixed before any claims were labeled, so every human label is evaluation data for it; we also confirm it on held-out submissions drawn after all development work (two annotators, $\kappa=0.89$ on 48 double-labeled claims). Across all labeled data, 78 of the 85 claims it certified were judged traceable (92\%, exact 95\% CI 84--97\%), and 27 of 29 on the held-out sample, and it certified about a quarter of all evaluated claims (23\% coverage). Its recall is deliberately limited, and this matters for how it is used: among the claims it did not certify, a majority were nevertheless judged traceable by the annotators (57\% in the confirmation sample and 72\% in the development sample). Treating ``not verified'' as ``false'' would therefore wrongly reject more claims than it correctly flags. The checker is a conservative certifier: a positive result is strong evidence of provenance, while a negative result routes a claim to human review rather than counting as evidence of misconduct. Certification establishes that a number is supported by the recorded artifacts, not that the underlying experiment is scientifically sound.

\paragraph{Contributions} This paper makes four contributions.
\begin{enumerate}
  \item \textbf{An empirical characterization of agent integrity failures.} Using the official judgments, human labels, and an independent checker, we show that, in this benchmark, unsupported numbers are mostly misattributed rather than absent from the agent's files (Sections~\ref{sec:res-judge}--\ref{sec:res-labels}).
  \item \textbf{A validated claim-to-evidence checker.} We introduce a deterministic, model-free method that traces numerical claims to execution artifacts, with per-claim decoy calibration, and validate it with a development/confirmation design, two annotators, and frozen operating points (Sections~\ref{sec:checker} and~\ref{sec:res-checker}).
  \item \textbf{Benchmark-wide measurements.} Applied to 99 of the 135 submissions, the checker shows large differences between agents in how traceable their reported numbers are, weakly associated with the official judge's integrity scores but revealing at least one agent whose headline numbers pass the judge while most of its numbers cannot be traced (Section~\ref{sec:res-bench}).
  \item \textbf{A judge-robustness analysis.} On 30 submissions, a second judge configuration orders the agents' vendor groups as the official judge does but is significantly more lenient, so absolute pass rates depend on the judge configuration; the sample is too small to establish benchmark-wide ranking stability or to isolate vendor effects (Section~\ref{sec:res-judge2}).
\end{enumerate}

For managers and analysts who consume AI-produced research, the practical implication is direct: the right question about an agent's report is not ``does this number appear in the outputs?'' but ``which run produced it?''---and that question can be answered automatically for a useful fraction of claims. All code, labels, and derived data are released.\ifanon\footnote{Repository link withheld for double-blind review.}\else\footnote{\url{https://github.com/mos3b-faqeeh/agent-integrity-analysis}}\fi

The rest of the paper is organized as follows. Section~\ref{sec:related} reviews related work. Section~\ref{sec:data} describes the benchmark and data. Section~\ref{sec:methods} presents the methods, including the checker and the labeling protocol. Section~\ref{sec:results} reports the results, Section~\ref{sec:discussion} discusses implications and limitations, and Section~\ref{sec:conclusion} concludes.

\section{Related work}\label{sec:related}

This paper connects five literatures that have so far developed largely in parallel: the evaluation of AI research agents; studies of how such agents fail, and in particular how they misrepresent their own work; the attribution and verification of generated text; the checking of reported numbers in science; and the use of LLMs as judges. A sixth, the information-systems literature on data quality, provenance, and trust in automation, supplies the frame for treating agent reports as data products whose fitness for use must be established before they inform decisions. We review each in turn and close with a structured comparison (Table~\ref{tab:positioning}) that locates our contribution.

\subsection{From coding assistants to research agents}\label{sec:rw-agents}

Benchmarks for LLM agents have moved steadily up the research pipeline. Early agent benchmarks measured whether agents could resolve software issues \citep{jimenez2024swebench} or run machine-learning experiments in sandboxed environments \citep{huang2024mlagentbench}. Later work evaluated end-to-end machine-learning engineering \citep{chan2025mlebench}, compared agents against human experts on open-ended research-engineering tasks \citep{wijk2024rebench}, and assessed data-driven discovery across scientific disciplines \citep{chen2025scienceagentbench,jansen2024discoveryworld}. A parallel line examines whether agents can reproduce published work, either computationally from released code \citep{siegel2024corebench} or from the paper alone \citep{starace2025paperbench}. At the generative end, LLMs have been shown to propose research ideas that experts judge more novel, though less feasible, than their own \citep{si2025ideas}, and systems such as the AI Scientist \citep{lu2024aiscientist,yamada2025aiscientistv2} go further, producing complete manuscripts with experiments, figures, and reviews.

Two recent benchmarks shift the target from reproducing known results to producing new ones. MLR-Bench evaluates agents on open-ended machine-learning research with an LLM judge \citep{chen2025mlrbench}. \emph{Into the Unknown} \citep{kim2026unknown} asks agents to write follow-up papers to recent work in quantum computing, plasma physics, computational chemistry, materials science, and scientific machine learning, and is distinctive in gating every scientific assessment behind an integrity audit. Because it releases the full file system of every run alongside the judge's node-level verdicts, it is also the first benchmark on which the provenance of each reported number can be examined directly. This is the property our study exploits.

\subsection{How research agents misrepresent their work}\label{sec:rw-integrity}

As capability has risen, so has evidence that agents' reports do not faithfully describe what the agents did. An independent evaluation of the AI Scientist found manuscripts containing hallucinated numerical results and experiments that did not support their stated conclusions \citep{beel2025sakana}. MLR-Bench reports that coding agents produce fabricated or invalidated experimental results in about 80\% of the cases it inspected \citep{chen2025mlrbench}, and the \emph{Into the Unknown} authors attribute most of their 114 integrity failures to claimed work that was never done or to numbers that contradict the recorded evidence \citep{kim2026unknown}. Two 2026 case studies point to selective rather than invented reporting. In InnovationEval, agents ran many near-identical training runs, submitted the best result per dataset, and disclosed the selection only in passing or not at all, although their own transcripts show awareness of the practice; once out-of-scope selection was removed, little of the claimed improvement remained \citep{owen2026innovationeval}. In a shadow evaluation on unpublished conference submissions, agents handled the engineering competently but made little scientific progress, and both resulting papers were rejected by the original authors \citep{kirgis2026openended}.

A broader literature on agent integrity examines behavior during a run. Agents tamper with evaluation code or leak held-out data when given the opportunity \citep{atinafu2026rewardhacking}, exploit test cases when tasks are made impossible \citep{zhong2026impossiblebench}, and pass benchmarks through exploits that the benchmark designers did not anticipate \citep{luo2026unearned}. Language models can strategically misreport their own actions under pressure \citep{scheurer2023deceive}, and honesty---whether a model states what it believes---can be measured separately from accuracy \citep{ren2025mask}. Even without incentives to deceive, models' stated reasoning need not reflect the process that produced their answers \citep{turpin2023unfaithful,lanham2023faithfulness}, and research agents may defend mechanisms that their own data contradict \citep{eulig2026cawm}.

Two gaps run through this body of work. First, the evidence that agent reports are unreliable comes almost entirely from expert review or a single LLM judge, validated, where at all, at the level of whole submissions; we know of no study that checks integrity verdicts claim by claim against blinded human labels, or tests them under a second, independently configured judge. Second, the studies do not separate two failure modes with different causes and remedies: \emph{fabrication}, where a reported number was never produced, and \emph{misattribution}, where it was produced but is reported for the wrong experiment, metric, or run. The selective-reporting evidence from InnovationEval suggests that the second may be common, but no study has measured the split. Our work does so at the level of individual numerical claims.

\subsection{Attribution and provenance of generated content}\label{sec:rw-attribution}

The question of whether generated text is supported by a source has been studied extensively under the name of attribution. \citet{rashkin2023ais} formalize when a statement is ``attributable to an identified source'' and show that human judgments of attribution can be made reliably. Systems that generate text with citations are evaluated by citation precision and recall \citep{gao2023alce}, and an audit of commercial generative search engines found that only about half of generated sentences were fully supported by their citations \citep{liu2023verifiability}. Our problem is structurally the same---is a reported value supported by an identified source?---but the sources differ. Attribution research assumes a small set of retrieved passages written for humans; we face thousands of machine-written result entries, many containing the same value for different quantities. The difficulty is less in judging support than in establishing \emph{which} entry a claim refers to.

Closest to our approach, a recent survey of AI-scientist systems identifies verification as the field's binding constraint, noting that released systems seldom expose the seeds, traces, and intermediate outputs needed to check a result \citep{ding2026survey}, and \citet{yin2026observability} propose that scientific agents emit artifact-centered, claim-aware event traces, sketching a checker that would compare a manuscript's numbers with experiment logs. Their contribution is a design specification: the checker is described but not implemented or evaluated. We implement the check, validate it against blinded human labels on held-out data, and show that it works on agent outputs as they are, without requiring agents to emit special traces. Our work also builds on the data-provenance tradition in databases and e-science \citep{simmhan2005provenance,herschel2017provenance}, its standardization in the W3C PROV model \citep{moreau2013prov}, and documentation practices for machine-learning artifacts \citep{gebru2021datasheets}, all of which treat the derivation of a data item as a first-class property that must be recorded to be trusted.

\subsection{Checking reported numbers and claims}\label{sec:rw-numbers}

Automated checks of reported numbers predate LLMs. In psychology, recomputing $p$-values from reported test statistics revealed widespread inconsistencies \citep{bakker2011errors}; the statcheck tool scaled this to roughly 250{,}000 $p$-values and found at least one inconsistency in about half of the papers examined \citep{nuijten2016statcheck}. The GRIM test flags reported means that are impossible given the sample size \citep{brown2017grim}. Studies of reproducibility in AI and computational science show how rarely published results can be regenerated from the information provided, and call for code, data, and workflow disclosure \citep{gundersen2018reproducibility,pineau2021reproducibility,stodden2016reproducibility}. LLMs now contribute to the scientific record directly---their use in writing scientific papers has risen sharply \citep{liang2024mapping}---and their tendency to fabricate details, from citations \citep{walters2023citations} to factual content more broadly \citep{huang2025hallucination}, is well documented.

Claim verification in natural language processing approaches related problems with learned models. Benchmarks pair claims with evidence from encyclopedias \citep{thorne2018fever}, scientific abstracts \citep{wadden2020scifact}, and tables \citep{chen2020tabfact,lu2023scitab}, and recent work targets numerical claims specifically \citep{venktesh2024quantemp}. These settings, however, assume curated evidence and model-based judgments of support.

The consistency checks test whether reported numbers are \emph{internally} coherent; the verification benchmarks test whether claims agree with \emph{external} reference text. Neither tests whether a number was produced by the run that supposedly produced it, because in conventional publishing the run's outputs are unavailable. Agent benchmarks remove this obstacle, and we argue that they also change what verification should mean: because misattributed numbers exist in the outputs, a check that confirms a value's presence passes exactly the errors that matter most. We therefore design a deterministic, provenance-scoring checker rather than a learned verifier, which would itself require verification, and we evaluate it on precision and on the cost of its non-verifications, not on presence alone.

\subsection{LLMs as judges of integrity}\label{sec:rw-judges}

LLM judges are now standard in evaluation. They agree with human preferences at levels comparable to inter-human agreement on many tasks \citep{zheng2023judge,chiang2023humaneval,liu2023geval}, but they are sensitive to the order in which candidates are presented \citep{wang2024notfair} and can recognize and prefer their own generations \citep{panickssery2024selfpref}; a recent survey catalogs these and other biases \citep{gu2024judgesurvey}. Cross-model review can surface errors that a same-model reviewer misses, though reviewer capability matters as much as independence \citep{song2026crossmodel}. Most of this evidence concerns judgments of quality or preference on short outputs. Integrity auditing is different in kind: the judge must read a long report, inspect a large file system, and decide whether specific statements are supported. Whether such verdicts are stable across repeated runs, and whether they depend on the judge's vendor when that vendor also builds some of the agents under evaluation, has not been tested. We examine both on a sample of submissions, re-grading them with a second judge configuration built on the benchmark's own grading pipeline.

\subsection{Data quality, provenance, and trust in AI-produced analyses}\label{sec:rw-is}

In information systems, data quality is defined by fitness for use: accuracy is one dimension among several, alongside believability, traceability, and interpretability \citep{wang1996dataquality,pipino2002dataquality}. Managing AI in organizations raises distinct challenges because AI systems act with increasing autonomy and their outputs are increasingly inscrutable to the people who rely on them \citep{berente2021managing}. Reliance on automated outputs is shaped by trust calibration: users over-trust automation that appears competent \citep{parasuraman2010complacency,lee2004trust} and abandon algorithms after seeing them err \citep{dietvorst2015aversion}. Recent work in this journal examines the trustworthiness of AI-generated content and the challenges of human--AI interaction in organizations \citep{li2026aigctrust,zhou2026humanai}. A report written by a research agent is a data product of exactly this kind. Its readers cannot inspect how each number was derived, and both failure modes of trust apply: accepting misattributed numbers, or discarding sound work because some numbers could not be checked. Our results speak to both. Value presence is a poor proxy for provenance, so naive checks invite over-trust; and most numbers a conservative checker cannot verify are nonetheless traceable, so treating non-verification as falsity invites unwarranted rejection.

\subsection{Positioning}\label{sec:rw-positioning}

Table~\ref{tab:positioning} compares representative approaches along the dimensions that define our contribution. Existing number checkers work within a paper; claim-verification benchmarks and attribution metrics test text against external or retrieved evidence; agent studies rely on expert or LLM judgment of whole submissions; and the one proposal that targets execution artifacts has not been built. To our knowledge, this paper is the first to (i) measure, at the level of individual reported numbers, how often agent integrity failures are misattribution rather than fabrication; (ii) implement and validate, on held-out data with blinded human labels, a deterministic check of numerical provenance against execution artifacts; (iii) quantify how often failure to verify coincides with a traceable claim; and (iv) test the robustness of a benchmark's integrity verdicts to the judge's vendor.

\begin{table}[t]
\centering
\caption{Positioning of this work relative to representative approaches. \textit{Evidence}: what a claim is checked against. \textit{Separates}: whether misattribution (value exists, wrong source) is distinguished from fabrication (value absent). \textit{Human-validated}: checked against human labels on held-out data. \textit{Judge robustness}: whether conclusions are tested under a second, independent judge. \checkmark{} = yes; -- = no or not applicable; (\checkmark) = partially.}
\label{tab:positioning}
\footnotesize
\setlength{\tabcolsep}{4pt}
\begin{tabular}{@{}>{\raggedright\arraybackslash}p{3.9cm}>{\raggedright\arraybackslash}p{2.6cm}>{\raggedright\arraybackslash}p{1.6cm}ccc@{}}
\toprule
Approach & Evidence & Unit & Separates & \shortstack{Human-\\validated} & \shortstack{Judge\\robustness}\\
\midrule
statcheck, GRIM \citep{nuijten2016statcheck,brown2017grim} & Paper itself & Statistic & -- & \checkmark & -- \\
FEVER, SciFact, SciTab, QuanTemp & Curated text or tables & Claim & -- & \checkmark & -- \\
Attribution metrics \citep{rashkin2023ais,gao2023alce,liu2023verifiability} & Retrieved passages & Sentence & -- & \checkmark & -- \\
MLR-Bench judge \citep{chen2025mlrbench} & Paper, code, logs (LLM) & Submission & -- & (\checkmark) & -- \\
\emph{Into the Unknown} judge \citep{kim2026unknown} & File system (LLM) & Rubric node & -- & -- & -- \\
InnovationEval \citep{owen2026innovationeval} & Transcripts (expert) & Submission & (\checkmark) & -- & -- \\
Claim-aware traces \citep{yin2026observability} & Emitted traces & Claim & -- & -- & -- \\
\textbf{This paper} & \textbf{Execution artifacts} & \textbf{Numeric claim} & \checkmark & \checkmark & \checkmark \\
\bottomrule
\end{tabular}
\end{table}

\section{Benchmark and data}\label{sec:data}

\subsection{The \emph{Into the Unknown} benchmark}

\emph{Into the Unknown} \citep{kim2026unknown} asks AI agents to write a follow-up research paper to each of fifteen recent papers in the computational and physical sciences. Each task supplies the original paper, a task specification, and a computing environment; the agent must formulate a direction, implement and run experiments, and submit a paper with its full working directory. Nine agents from three vendor groups each attempted all fifteen tasks, giving 135 submissions (Table~\ref{tab:agents}).

Each submission was graded by an LLM judge (GPT-6-Astra) against a structured rubric. The rubric is a tree of binary criteria (\emph{nodes}) organized into three sequential gates: G1, research integrity (for example, whether reported values trace to recorded outputs and whether the described method matches the code); G2, completeness of the deliverables; and G3, whether the work meets the bar a domain scientist would expect. Each node receives a status of pass, fail, or undetermined, together with a written justification and pointers to evidence. The judge evaluated every submission four times, and the benchmark aggregates the repeats into a consensus verdict. A submission passes a gate only if it passes all counted nodes in that gate. No submission passed all three gates; 21 passed G1 (Table~\ref{tab:agents}).

\subsection{Released artifacts}

The benchmark release contains, for every submission, (i) the four judge outputs and the consensus verdict, with node-level statuses and justifications; (ii) the agent's working directory as the judge saw it, including the paper's \LaTeX{} source and PDF, code, configuration files, logs, and all result files the agent wrote; and (iii) for some harnesses, the agent's full interaction trajectory. We use the release at a fixed snapshot (repository commit \texttt{612cdca}). Our analyses rely only on these public artifacts; we do not re-run any agent.

Three properties of the release make claim-level provenance analysis possible. The working directories are complete, so every result file the agent produced is available. The papers are written in \LaTeX{}, so numbers and their surrounding text can be extracted reliably, including from tables and from macros that agents use to insert computed values. And the judge's node-level justifications are released, so the reasons behind each integrity failure can be analyzed directly.

\begin{table}[t]
\centering
\caption{Agents in the benchmark. G1: submissions (of 15) passing the integrity gate under the official consensus verdict. Hours: median wall-clock duration of a run. $R_\text{int}$: mean share of integrity nodes passed (\%).}
\label{tab:agents}
\footnotesize
\begin{tabular}{@{}llrrr@{}}
\toprule
Agent & Vendor group & G1 passes & Hours & $R_\text{int}$\\
\midrule
astra-6 & OpenAI & 11 & 1.3 & 97.8\\
sol-5.6 & OpenAI & 6 & 2.9 & 88.7\\
opus-5.5 & Anthropic & 4 & 5.3 & 90.4\\
opus-5 & Anthropic & 0 & 7.5 & 59.5\\
opus-4.8 & Anthropic & 0 & 4.8 & 65.3\\
fable-5 & Anthropic & 0 & 7.0 & 66.5\\
deepseek-v4-pro & Open-weight & 0 & 19.5 & 64.6\\
kimi-k3 & Open-weight & 0 & 5.6 & 57.4\\
glm-5.3 & Open-weight & 0 & 13.9 & 54.4\\
\midrule
All & & 21 & & \\
\bottomrule
\end{tabular}
\end{table}

\section{Methods}\label{sec:methods}

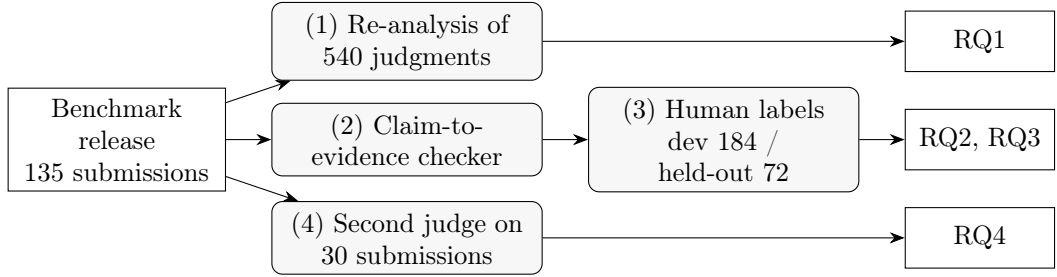
\begin{figure}[t]
\centering
\begin{tikzpicture}[node distance=4mm and 6mm, font=\footnotesize,
  box/.style={draw, rounded corners, align=center, minimum height=9mm, text width=3.3cm, fill=gray!6},
  data/.style={draw, align=center, minimum height=8mm, text width=2.6cm},
  arr/.style={-{Stealth[length=2mm]}}]
\node[data] (rel) {Benchmark release\\ 135 submissions};
\node[box, right=of rel, yshift=13mm] (re) {(1) Re-analysis of\\ 540 judgments};
\node[box, right=of rel] (ck) {(2) Claim-to-evidence checker};
\node[box, right=of rel, yshift=-13mm] (j2) {(4) Second judge on\\ 30 submissions};
\node[box, right=of ck] (lab) {(3) Human labels\\ dev 184 / held-out 72};

\node[data, right=of lab, text width=1.7cm] (o2) {RQ2, RQ3};
\node[data, text width=1.7cm] (o1) at (o2 |- re) {RQ1};
\node[data, text width=1.7cm] (o3) at (o2 |- j2) {RQ4};
\draw[arr] (rel) -- (re); \draw[arr] (rel) -- (ck); \draw[arr] (rel) -- (j2);
\draw[arr] (ck) -- (lab); \draw[arr] (re) -- (o1); \draw[arr] (lab) -- (o2); \draw[arr] (j2) -- (o3);
\end{tikzpicture}
\caption{Overview of the study. The checker is developed on a development sample and evaluated, with its rules frozen, on held-out submissions against blinded human labels.}
\label{fig:overview}
\end{figure}

Our analysis has four components (Fig.~\ref{fig:overview}): a re-analysis of the official judgments (Section~\ref{sec:reanalysis}); a deterministic claim-to-evidence checker (Section~\ref{sec:checker}); a human labeling study that provides the reference standard for the checker and for the misattribution--fabrication distinction (Section~\ref{sec:labels}); and a re-grading of a stratified sample by a second judge (Section~\ref{sec:judge2}).

\subsection{Re-analysis of the official judgments}\label{sec:reanalysis}

From the 540 released judge outputs we extract every counted G1 node: 2,376 node verdicts at the consensus level, each with a status, a confidence, a justification, and evidence pointers. Nodes belong to named rubric families (for example, \emph{code--paper alignment}, \emph{number traceability}, \emph{test-set integrity}, \emph{no hard-coded values}). We report failure rates by family, which separates criteria that check whether the write-up matches the work from criteria that check for overt misconduct. To assess the judge's self-consistency, we compute the share of nodes on which all four repeats agree.

We also test whether integrity outcomes are explained by how long an agent ran. Long runs plausibly lead to messier records and more opportunities for misreporting. For submission $i$ by agent $a$ on task $t$, let $R_{\text{int},i}$ be the share of counted G1 nodes passed and $h_i$ the run duration in hours. We estimate
\begin{equation}
R_{\text{int},i} = \beta_0 + \beta_1 \log h_i + \alpha_{a(i)} + \gamma_{t(i)} + \varepsilon_i,
\label{eq:fe}
\end{equation}
where $\alpha_a$ and $\gamma_t$ are agent and task fixed effects, with heteroskedasticity-robust (HC1) standard errors. A coefficient $\beta_1$ that is large without the fixed effects but vanishes with them indicates that run length proxies for agent identity rather than causing integrity failures.

\subsection{Claim-to-evidence checker}\label{sec:checker}

The checker takes a submission's paper and working directory and assigns each numerical claim one of three verdicts: \emph{certified} (the claim's value is found in a result file whose context identifies it, beyond what chance would produce), \emph{unattributed} (the value is found but no source is identified), or \emph{untraced} (the value does not occur in any result file at the reported precision). It is deterministic and uses no learned model, so its behavior can be audited and reproduced exactly.

\paragraph{Design principle} Existing approaches ask different questions of a reported number. Consistency checks such as statcheck and GRIM ask whether the number is \emph{possible} given other numbers in the paper. Value matching asks whether the number \emph{appears} somewhere in the outputs. Claim-verification models ask whether retrieved text \emph{supports} a statement. Our checker asks a narrower question that none of these answers: does the number \emph{belong to the specific result the paper says it describes}? A value can be possible, present, and supported by some text while still being attached to the wrong experiment, as the worked example below shows: the value 0.1000 appears 24 times in one agent's files, but the entry that best matches it is a configuration setting, not the reported result. Answering the narrower question requires two ingredients that value matching lacks: a measure of whether the evidence entry's context identifies the claimed quantity, and a per-claim test of whether that identification beats chance.

\paragraph{Claims} The checker reads the paper's main \LaTeX{} file, expands \verb|\input| and \verb|\include| commands and numeric macro definitions (agents frequently insert computed values through macros), and removes the bibliography. It extracts every number with a decimal point, percent sign, or scientific exponent; integers are excluded because they are mostly counts, indices, and years. Each claim records its value, its number of decimal places, whether it is a percentage, and its number of significant digits. Its context is the surrounding 160 characters for numbers in prose, and the row label, column header, and caption for numbers in tables. We evaluate claims with at least two significant digits, since one-digit values match too many unrelated numbers to be informative.

\paragraph{Evidence} Evidence consists of machine-written result files only: JSON, JSONL, CSV, TSV, log, output, and plain-text files in the working directory. Text written for humans is excluded: the paper's own source and any rendered copy of it, \LaTeX{} build logs, code, and Markdown notes. A number reported in the agent's prose or code is not evidence that an experiment produced it. Every numeric value in an evidence file becomes an evidence entry with a context string: for JSON, the path of keys leading to the value; for CSV, the column header and the row's leading fields; for logs, the line itself. Entries from bulk data---JSON arrays of more than 20 numbers, CSV or JSONL files with more than 200 rows, and log lines containing more than 10 numbers---are flagged, because a reported summary statistic rarely comes from a raw data dump.

\paragraph{Notation} A claim is a tuple $c=(v_c, d_c, \pi_c, C_c)$: its value, number of decimal places, a percentage indicator, and the set of content tokens in its context. An evidence entry is a tuple $e=(u_e, E_e, b_e)$: its value, the set of content tokens in its key path, line, and file path, and a bulk indicator. Let $\mathcal{E}$ be the set of entries extracted from a submission.

\paragraph{Level 1: value match} A claim matches an entry when the entry's value rounds to the claim at the claim's reported precision:
\begin{equation}
M(c,e) = \mathbb{1}\Bigl[\,\min_{\lambda\in\Lambda_c} \bigl|\lambda u_e - v_c\bigr| \le \delta_c \Bigr], \qquad \delta_c = \tfrac{1}{2}\,10^{-d_c}\,(1+10^{-9}),
\label{eq:match}
\end{equation}
where $\Lambda_c=\{1,100\}$ if $\pi_c=1$ (a percentage, so 8.04\% matches 0.0804) and $\Lambda_c=\{1\}$ otherwise; the factor $1+10^{-9}$ absorbs floating-point error. The matched set is $\mathcal{M}(c)=\{e\in\mathcal{E}: M(c,e)=1\}$, and Level 1 accepts $c$ whenever $\mathcal{M}(c)\neq\emptyset$. Level 1 is the obvious automated test---does the number appear anywhere?---and we report it as a baseline.

\paragraph{Level 2: provenance score} Among non-bulk matches $\mathcal{M}^{*}(c)=\{e\in\mathcal{M}(c): b_e=0\}$, the checker scores how well each entry's context identifies the claim:
\begin{equation}
s(c,e) = \frac{|C_c\cap E_e|}{\sqrt{|C_c\cup E_e|}\,\sqrt{|C_c|}}, \qquad S(c) = \max_{e\in\mathcal{M}^{*}(c)} s(c,e),
\label{eq:score}
\end{equation}
with $S(c)=0$ when $\mathcal{M}^{*}(c)=\emptyset$. The score is a cosine-like overlap that rewards evidence keys sharing words with the sentence or table cell in which the number appears, and the normalization by $|C_c\cup E_e|$ penalizes long, generic contexts that overlap with everything.

\paragraph{Per-claim decoy calibration} A high score can arise by chance when a file contains many similar numbers with similar keys. To guard against this, the checker repeats the scoring for decoy claims $c^{(k)}$ that are identical to $c$ except for their value,
\begin{equation}
v_{c^{(k)}} = v_c + k\,10^{-d_c}, \qquad k\in K=\{\pm2,\pm4,\pm6,\pm8,\pm10\},
\label{eq:decoy}
\end{equation}
keeping only decoys with the same sign as $v_c$ (for 0.4583 the decoys are 0.4573, \ldots, 0.4593). The claim's calibration $p$-value is the rank of its own score among the decoys' scores,
\begin{equation}
p(c) = \frac{1+\sum_{k\in K}\mathbb{1}\bigl[S(c^{(k)})\ge S(c)\bigr]}{1+|K|},
\label{eq:pvalue}
\end{equation}
and the verdict is
\begin{equation}
V(c)=\begin{cases}
\text{certified} & \text{if } S(c)\ge\tau \text{ and } p(c)\le\rho,\\
\text{untraced} & \text{if } \mathcal{M}^{*}(c)=\emptyset,\\
\text{unattributed} & \text{otherwise.}
\end{cases}
\label{eq:verdict}
\end{equation}
With ten decoys, $p(c)\le 0.1$ means the true value outscores every decoy. A further decoy at offset $k=+7$, never used in Eq.~\eqref{eq:pvalue}, measures how often a value that is certainly wrong would be certified (the \emph{decoy acceptance rate}). Algorithm~\ref{alg:checker} summarizes the procedure.

\begin{algorithm}[t]
\caption{Claim-to-evidence checker for one submission}
\label{alg:checker}
\footnotesize
\begin{algorithmic}[1]
\Require paper source $P$, working directory $W$, threshold $\tau$, calibration level $\rho$
\State $\mathcal{C}\gets$ \Call{ExtractClaims}{$P$} \Comment{expand macros; prose window or row\,|\,column\,|\,caption}
\State $\mathcal{E}\gets$ \Call{IndexEvidence}{$W$} \Comment{result files only; flag bulk entries}
\State build a sorted value index over $\mathcal{E}$ \Comment{range queries for Eq.~\eqref{eq:match}}
\For{each claim $c\in\mathcal{C}$}
  \State $\mathcal{M}(c)\gets\{e: M(c,e)=1\}$;\ \ $\mathcal{M}^{*}(c)\gets\{e\in\mathcal{M}(c): b_e=0\}$
  \If{$\mathcal{M}^{*}(c)=\emptyset$} \State $V(c)\gets$ untraced; \textbf{continue} \EndIf
  \State $S(c)\gets\max_{e\in\mathcal{M}^{*}(c)} s(c,e)$ \Comment{Eq.~\eqref{eq:score}}
  \For{each $k\in K$} \State $S(c^{(k)})\gets$ score of decoy $c^{(k)}$ \Comment{Eq.~\eqref{eq:decoy}} \EndFor
  \State $p(c)\gets$ Eq.~\eqref{eq:pvalue}
  \State $V(c)\gets$ certified \textbf{if} $S(c)\ge\tau$ \textbf{and} $p(c)\le\rho$ \textbf{else} unattributed
\EndFor
\State \Return $\{V(c)\}_{c\in\mathcal{C}}$ with best-matching entries for audit
\end{algorithmic}
\end{algorithm}

\paragraph{Operating points} The strict operating point ($\tau=0.12$, $\rho=0.1$) was fixed before any human labels were collected. One change was made to the checker afterwards: during development labeling we noticed that the agent's own rendered paper text and \LaTeX{} build logs were being indexed as evidence, and excluded them, as the evidence definition above requires. This correction left the development test-split results unchanged and was made before the confirmation sample was drawn. After the development labels were complete, we examined relaxed variants on the development tuning split only, and froze two of them---\emph{balanced} ($\tau=0$, $\rho=0.1$) and \emph{broad} ($\tau=0$, $\rho=0.2$)---before the held-out confirmation sample was drawn. Strict is the primary operating point throughout.

\paragraph{Worked example} Two claims from the development sample illustrate the difference between the levels. In ``the high-density half-chain entropy rises \ldots\ to 0.8749 bits at 4.0'', the value 0.8749 matches two evidence entries; the best is \texttt{"half\_chain\_entropy\_bits": 0.87487} in an output file, with score 0.38, and no decoy scores as high, so the claim is certified. In ``the matched raw-sensor encoder obtains \ldots\ 0.1000'', the value 0.1000 matches 24 entries, so Level 1 accepts it; but the best-scoring entry is a configuration setting (\texttt{train\_profiles g3 base}) with score 0.11, below the threshold, and the claim is left unattributed. Human annotators labeled the first claim traceable and the second wrong-source: the number is present, but not as the result the sentence describes.

\subsection{Human labeling}\label{sec:labels}

\paragraph{Labels} Annotators assign each claim one of four labels. \emph{Traceable}: a result entry gives this value for the quantity the sentence or table cell describes (same experiment, metric, and model; rounding allowed). \emph{Wrong source}: the value occurs only for a different experiment, metric, run, dataset, or statistic (for example, a standard deviation where a mean is reported). \emph{Not found}: no entry in the available artifacts supports the value. \emph{Not a claim}: the number is a setting rather than a result (a grid value, tolerance, threshold, confidence level, or chosen parameter) or an extraction error. We use ``not found'' strictly in the sense of absence from the available artifacts. It is not a finding of fabrication, since a value may have been computed and not saved.

\paragraph{Presentation} For each claim, annotators see the paper context with the number highlighted and the three best-matching evidence entries (value, file, and context), ordered so that non-bulk entries come first; they may also consult other files. They never see the checker's verdict, score, or $p$-value. Verdicts were stored in a separate file that was not opened during labeling.

\paragraph{Development sample} We drew ten submissions spanning all three vendor groups (one contained no extractable claims) and sampled up to 22 claims per submission with at least two significant digits, stratified so that about 40\% came from claims the strict checker certified (to estimate precision with useful counts) and the rest from the remainder. Population estimates reweight each claim by the inverse of its sampling probability within its submission and stratum. The first ten claims served as annotator practice and are excluded, leaving 184 claims, which were split alternately into tuning and test halves. Labeling rules that emerged on the development sample (for instance, that values computed by hand from settings count as settings) were fixed before the confirmation sample was drawn. To measure self-consistency, the first annotator relabeled 20 earlier claims, renamed and shuffled, without access to the original labels.

\paragraph{Confirmation sample} After the operating points were frozen, we drew one previously unused submission per agent (nine in total) and sampled nine claims from each with the same stratification. One submission (astra-6) was dropped before any of its claims were viewed, because the checker had not finished processing it, leaving 72 claims from eight submissions. The first annotator labeled all 72. The second annotator, a co-author who had not taken part in development labeling, labeled a subset of 48 (six per submission) after reading a written guide with three examples taken from the development sample, blind to the first annotator's labels and to the checker; the first annotator was present during this session but did not comment on any claim. Disagreements were not adjudicated: we report results against the first annotator's labels and, as a sensitivity analysis, against the claims on which both annotators agree. During confirmation labeling, four of the first annotator's labels conflicted with the written rules (for instance, a grid spacing labeled traceable); after this was pointed out in a consistency review against the rules, performed with an AI assistant, the annotator revised them. We report agreement both before and after these revisions.

\paragraph{Metrics} Table~\ref{tab:metrics} defines the quantities we report. We treat \emph{traceable} as the positive class and exclude claims labeled \emph{not a claim}. Because the samples over-represent certified claims, population estimates weight each labeled claim $j$ by the inverse of its sampling probability within its submission $m$ and stratum $g$ (certified or not):
\begin{equation}
w_j = \frac{N_{mg}}{n_{mg}},
\label{eq:weights}
\end{equation}
where $N_{mg}$ is the number of eligible claims and $n_{mg}$ the number sampled. With $y_j=1$ for a traceable claim and $\hat y_j=1$ for a certified one, weighted precision and recall are
\begin{equation}
\widehat{\mathrm{Prec}} = \frac{\sum_j w_j\,\hat y_j\,y_j}{\sum_j w_j\,\hat y_j}, \qquad
\widehat{\mathrm{Rec}} = \frac{\sum_j w_j\,\hat y_j\,y_j}{\sum_j w_j\,y_j},
\label{eq:precrec}
\end{equation}
and coverage is $\sum_j w_j\hat y_j/\sum_j w_j$. Confidence intervals are 95\% percentile intervals from a cluster bootstrap that resamples submissions (2,000 resamples), which accounts for claims from the same paper being correlated; for precision we also report exact Clopper--Pearson intervals on unweighted counts. Agreement between annotators is measured with Cohen's $\kappa$ \citep{cohen1960kappa},
\begin{equation}
\kappa = \frac{p_o - p_e}{1-p_e},
\label{eq:kappa}
\end{equation}
where $p_o$ is the observed proportion of agreement and $p_e$ the agreement expected from the annotators' marginal label frequencies, interpreted with conventional bands \citep{landis1977kappa}.

\begin{table}[t]
\centering
\caption{Quantities used to evaluate the checker. ``Real claims'' excludes claims labeled \emph{not a claim}.}
\label{tab:metrics}
\footnotesize
\begin{tabular}{@{}>{\raggedright\arraybackslash}p{3.0cm}>{\raggedright\arraybackslash}p{9.6cm}@{}}
\toprule
Quantity & Definition and interpretation\\
\midrule
Precision & Share of certified claims that annotators labeled traceable. How far a positive verdict can be trusted.\\
Recall & Share of traceable claims that the checker certifies. How much of the verifiable content it finds.\\
Coverage & Share of all real claims that the checker certifies. How much of a paper it settles automatically.\\
Traceable among uncertified & Share of claims the checker did not certify that annotators labeled traceable. How often a negative verdict would be a false accusation if read as one.\\
Decoy acceptance & Share of held-out decoy values (offset $+7$) that would be certified. A label-free estimate of false certification.\\
\bottomrule
\end{tabular}
\end{table}

\subsection{Baselines}\label{sec:baselines}

We compare the checker with two alternatives that a practitioner might reach for. The \emph{embedding baseline} keeps the checker's matching, bulk filter, and decoy test but replaces the token-overlap score of Eq.~\eqref{eq:score} with the cosine similarity between sentence embeddings (all-MiniLM-L6-v2) of the claim context and of each evidence entry's context and file path. It is run with and without the decoy test, and its threshold is tuned on the development tuning split to match the strict checker's precision there. The \emph{LLM verifier} gives a large language model exactly what the human annotators saw---the written labeling guide, the paper context, and the three best-matching evidence entries---and asks for one of the four labels, with no access to files, human labels, or checker verdicts; a \emph{traceable} answer counts as a certification. We ran it with Claude Opus 5.5 through the vendor's command-line interface, one single-turn call per claim in an empty working directory. All methods are compared on the 129 labeled real claims that were not used for any tuning: the development test split and the confirmation sample.

\subsection{Second judge}\label{sec:judge2}

To test whether the integrity verdicts depend on the judge, we re-graded a stratified sample of 30 submissions with a second judge configuration: Claude Opus 5.5 run in the benchmark's Claude Code judge runner. This runner reuses the official prompt, rubric, output schema, validator, and deterministic grading code. The judge model, the execution harness (a Claude Code session rather than a Codex thread), and the file-access restrictions all differ from the official configuration, so the comparison is between judge configurations and cannot isolate the effect of the judge's vendor. The sample contains three or four submissions per agent, including up to two G1 passers for each agent that had any, and was restricted to submissions whose archived working directory is at most 5 GB. Because of computational cost, the first two submissions were graded four times, the next two twice, and the remaining 26 once.

We compare the two judges node by node. Per-repeat agreement compares each second-judge repeat with each official repeat on the counted nodes both graded; gate agreement compares majority verdicts. To test for a vendor effect, let $\mathcal{N}_i$ be the counted G1 nodes graded by both judges for submission $i$ and $f^{J}_{in}=1$ if judge $J$ fails node $n$. The leniency difference is
\begin{equation}
d_i = \frac{1}{|\mathcal{N}_i|}\sum_{n\in\mathcal{N}_i}\bigl(f^{\text{second}}_{in}-f^{\text{official}}_{in}\bigr).
\label{eq:d}
\end{equation}
We test whether $d_i$ is non-zero overall (Wilcoxon signed-rank test) and whether it differs across vendor groups (Kruskal--Wallis test, and permutation tests for each group against the rest), using submissions as the unit of analysis. As a node-level check, we fit
\begin{equation}
\begin{split}
\operatorname{logit}\Pr\bigl(f^{J}_{in}=1\bigr) = {}& \beta_0 + \beta_1\,\mathbb{1}[J=\text{second}] + \sum_{v}\beta_{2v}\,\mathbb{1}[g_i=v]\\
&+ \sum_{v}\beta_{3v}\,\mathbb{1}[J=\text{second}]\,\mathbb{1}[g_i=v],
\end{split}
\label{eq:gee}
\end{equation}
where $g_i$ is the vendor group of the agent (open-weight as reference), estimated by generalized estimating equations with an exchangeable working correlation clustered by submission. A vendor effect would appear as interaction terms $\beta_{3v}$ that differ in sign or size across groups.

\section{Results}\label{sec:results}

\subsection{What the official judge fails (RQ1)}\label{sec:res-judge}

\begin{figure}[t]
\centering
\includegraphics[width=0.9\linewidth]{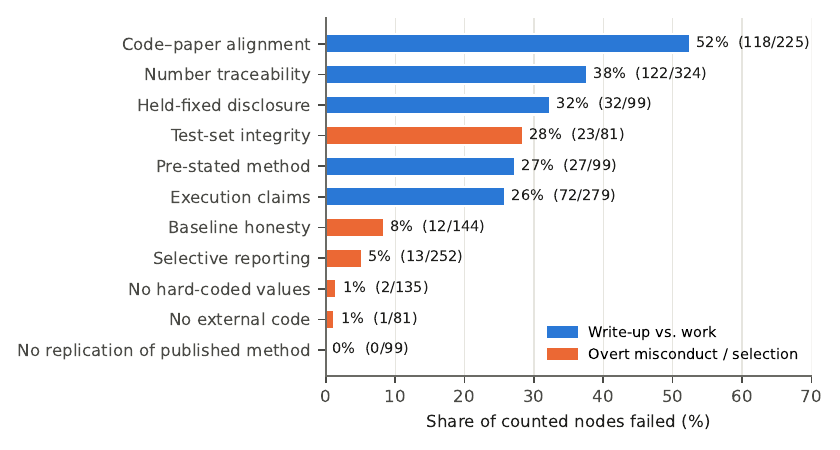}
\caption{Failure rates of integrity (G1) rubric families under the official judge's consensus verdicts, with failed and counted nodes in parentheses. Criteria comparing the write-up with the work (blue) fail far more often than criteria targeting overt misconduct or selection (orange).}
\label{fig:families}
\end{figure}

Of the 540 released judgments, 450 ended in an integrity failure, 86 in an incompleteness verdict, and 4 below the scientist's bar. Across the 2,376 counted G1 nodes at the consensus level, 469 failed, almost all (438) with high confidence. Fig.~\ref{fig:families} shows how these failures distribute across rubric families. Failures concentrate in criteria that compare the write-up with the work: code--paper alignment fails in 52\% of counted nodes and number traceability in 38\%, followed by disclosure of what was held fixed (32\%), adherence to the pre-stated method (27\%), and claims about what was executed (26\%). Criteria that target overt misconduct almost never fire: hard-coded values (1\%), use of external code (1\%), and replication of the published method instead of new work (0\%). Selective reporting (5\%) and baseline honesty (8\%) are also rare as recorded by the judge.

The judge's justifications show what these failures look like in practice. Typical examples state that a paper claims 400 training epochs where the logs show 320; claims ten seeds where nine were run; describes an ``identical schedule'' across conditions where the records show different schedules; or describes a model term that the final simulation never applies. In each case the paper describes work that resembles what was done, but not exactly what was done. Failure rates differ sharply between agents: code--paper alignment fails in 0\% of nodes for astra-6, 16\% for sol-5.6, and 28\% for opus-5.5, but in 56--84\% for the other six agents.

\paragraph{Judge self-consistency} The four repeats agree unanimously on 90.0\% of counted nodes (9,570 of 10,638; 89.5\% for G1), and the G1 verdict differs across repeats for 11 of the 135 submissions. The official verdicts are thus stable across repeated runs of the same judge, which makes the cross-judge comparison in Section~\ref{sec:res-judge2} interpretable.

\paragraph{Run length} Longer runs have lower integrity scores (Spearman $\rho=-0.37$, $p<10^{-4}$; median 2.3 hours for G1 passers against 9.2 hours for failures). The association disappears once agent and task are controlled for: in the fixed-effects regression, the coefficient on log hours is $-0.61$ ($p=0.59$; $R^2=0.77$), and within-agent correlations are all small and non-significant. Integrity differences are a property of the agent, not a by-product of how long it ran (Table~\ref{tab:fe} in \ref{app:tables}).

\subsection{Misattribution versus absence in human labels (RQ2)}\label{sec:res-labels}

Table~\ref{tab:labels} shows the label distribution. Among real claims (excluding settings and extraction errors), most numbers are traceable, as expected for samples that over-represent certified claims. The question for RQ2 is the composition of the remainder. Of the 53 unsupported claims across both samples, 47 are \emph{wrong source}---the value exists in the agent's files but for a different experiment, metric, run, or statistic---and only 6 are \emph{not found} in the available artifacts. Reweighting to the population of claims gives a wrong-source share of 81\%. In this benchmark, an unsupported number is usually a real number in the wrong place.

\begin{table}[t]
\centering
\caption{Human labels. Development: first annotator, ten submissions (after excluding practice items). Confirmation: first annotator, eight held-out submissions. ``Wrong source / unsupported'' is the share of wrong-source claims among wrong-source plus not-found claims, raw and reweighted to the population of claims.}
\label{tab:labels}
\footnotesize
\begin{tabular}{@{}lrrr@{}}
\toprule
Label & Development & Confirmation & Total\\
\midrule
Traceable & 112 & 46 & 158\\
Wrong source & 35 & 12 & 47\\
Not found & 4 & 2 & 6\\
Not a claim & 33 & 12 & 45\\
\midrule
Total labeled & 184 & 72 & 256\\
Wrong source / unsupported & 90\% & 86\% & 89\%\\
\quad population-weighted & & & 81\%\\
\bottomrule
\end{tabular}
\end{table}

The wrong-source claims fall into recurring patterns: a standard deviation reported as a mean; a value from the baseline model reported for the proposed one; a single seed's result reported as the average over seeds; a value from a different dataset or boundary condition; and, in a few cases, a matching number from an unrelated line such as a timer or a training-loss print. Table~\ref{tab:patterns} gives one example of each pattern from the labeled samples.

\begin{table}[t]
\centering
\caption{Recurring patterns among wrong-source claims, with one example of each from the labeled samples. In every case the reported value exists in the agent's files, but not as the quantity the paper describes.}
\label{tab:patterns}
\footnotesize
\begin{tabular}{@{}>{\raggedright\arraybackslash}p{2.4cm}>{\raggedright\arraybackslash}p{4.6cm}>{\raggedright\arraybackslash}p{5.4cm}@{}}
\toprule
Pattern & Reported in the paper & Where the value actually occurs\\
\midrule
Wrong statistic & ``0.045 vs.\ 0.026'' relative $L_2$ error of the proposed encoder & A standard deviation (\texttt{rel\_l2\_std} $=0.0453$) of a different ablation run\\
Wrong model or arm & Fidelity 0.9992 for the proposed method & \texttt{final\_fid} $=0.99920$ of the unconstrained baseline run\\
Wrong aggregation & Mean relative $L_2$ over three seeds & A single seed's value, or the absolute rather than relative error\\
Wrong condition & Error 0.144 for the raw-input control at coefficient (2,1) & \texttt{rel\_mean} $=0.1437$ of a different (learned-encoder) arm\\
Unrelated line & CNOT overhead multiplier 1.57 & A timing line, \texttt{one\_step 1.57 s}\\
Setting, not result & Encoder result 0.1000 & A training-profile configuration value\\
\bottomrule
\end{tabular}
\end{table}

The six not-found claims are heterogeneous: some quote values from prior literature, some appear to be quantities derived while writing and never saved, and some have no candidate source at all. From the artifacts alone, none can be classified as fabricated. The wrong-source pattern, by contrast, is consistent with reporting that has become detached from the record rather than with invented experiments.

\paragraph{Reliability of the labels} On the 48 confirmation claims labeled by both annotators, agreement is 93.8\% over the four labels (Cohen's $\kappa=0.89$) and 97.9\% on traceable versus not ($\kappa=0.96$), ``almost perfect'' by conventional bands \citep{landis1977kappa}. Using the first annotator's labels before the four rule-based revisions described in Section~\ref{sec:labels}, agreement is $\kappa=0.75$ (four labels) and $\kappa=0.83$ (binary); in all four cases the second annotator had independently chosen the revised label. The first annotator's blind re-labeling of 20 development claims agreed with the original labels in 75\% of cases ($\kappa=0.56$; 85\% and $\kappa=0.69$ for traceable versus not), with every change moving toward \emph{wrong source}. The traceable class in the development labels is therefore, if anything, generous.

\subsection{Can provenance be checked automatically? (RQ3)}\label{sec:res-checker}

\begin{figure}[t]
\centering
\includegraphics[width=0.62\linewidth]{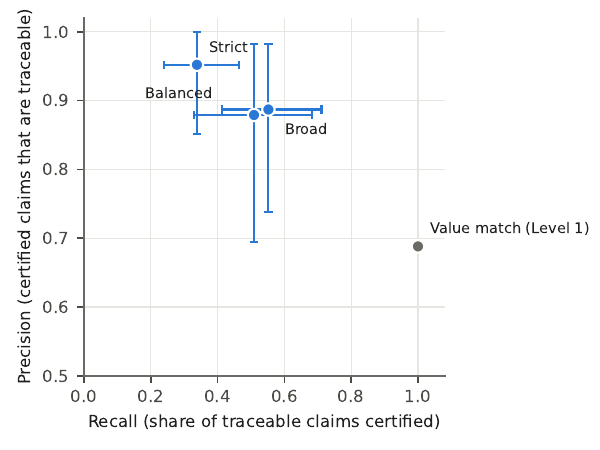}
\caption{Precision and recall of the checker on the held-out confirmation sample, with 95\% cluster-bootstrap intervals. Value matching (gray) finds every traceable claim but accepts most untraceable ones; the strict operating point trades recall for high precision.}
\label{fig:oppoints}
\end{figure}

Table~\ref{tab:checker} reports the checker's performance on the development test split and on the held-out confirmation sample; Fig.~\ref{fig:oppoints} plots the held-out results.

\paragraph{Value matching does not discriminate} Level~1 matching finds a value for essentially every traceable claim, but it also accepts most claims that are not traceable: on the held-out sample its specificity is 13\% (5\% on development), meaning it accepts 87--95\% of wrong-source and not-found claims. Its precision (0.69) is essentially the base rate of traceable claims, because it rejects almost nothing. The decoy test gives the same picture without labels: 88\% of the held-out decoy values, which are wrong by construction, match some entry at the reported precision.

\begin{table}[t]
\centering
\caption{Checker performance against the first annotator's labels (real claims only). Precision and recall are population-weighted with 95\% cluster-bootstrap intervals. Decoy acceptance is computed over the 1,351 development claims with at least two significant digits. Operating points were frozen before the confirmation sample was drawn.}
\label{tab:checker}
\footnotesize
\setlength{\tabcolsep}{3pt}
\resizebox{\linewidth}{!}{%
\begin{tabular}{@{}lccccc@{}}
\toprule
& \multicolumn{2}{c}{Development (test split, $n=69$)} & \multicolumn{2}{c}{Confirmation ($n=60$)} & Decoy\\
\cmidrule(lr){2-3}\cmidrule(lr){4-5}
Setting & Precision & Recall & Precision & Recall & acceptance\\
\midrule
Level 1 (value match) & 0.77 & 1.00 & 0.69 & 1.00 & 88\%\\
Strict & 0.97 [0.81, 1.00] & 0.39 [0.24, 0.54] & 0.95 [0.85, 1.00] & 0.34 [0.24, 0.47] & 1.3\%\\
Balanced & 0.84 [0.58, 0.96] & 0.49 [0.38, 0.84] & 0.88 [0.70, 0.98] & 0.51 [0.33, 0.68] & 3.8\%\\
Broad & 0.86 [0.61, 0.98] & 0.71 [0.58, 0.91] & 0.89 [0.74, 0.98] & 0.55 [0.41, 0.71] & 8.7\%\\
\bottomrule
\end{tabular}}
\end{table}

\paragraph{Provenance scoring gives a high-precision certifier} At the strict operating point, 27 of the 29 claims the checker certified on the held-out sample were labeled traceable (precision 0.93 unweighted, 0.95 population-weighted, cluster-bootstrap 95\% CI 0.85--1.00), and 0.94 against the claims on which both annotators agree. Because the held-out sample is small, Table~\ref{tab:precision} also reports exact binomial intervals and a pooled estimate. Pooling is legitimate for the strict operating point because its parameters were fixed before any claim was labeled, so no label influenced it. Across all 211 labeled real claims, the strict checker certified 85, of which 78 were traceable: a precision of 0.92 with an exact 95\% interval of 0.84--0.97. The held-out sample was deliberately small in real claims (60) because it over-samples certified claims to estimate precision efficiently; precision rests on the number of certified claims, not on the total. Label-free, the strict checker certifies 1.3\% of decoy values. Certification is therefore strong evidence that a reported number is supported by the result the paper says it is.

\begin{table}[t]
\centering
\caption{Precision of the strict checker with counts. Exact intervals are Clopper--Pearson 95\% intervals on the unweighted counts; they ignore clustering by submission and are reported as a conservative complement to the weighted, cluster-bootstrap estimates in Table~\ref{tab:checker}.}
\label{tab:precision}
\footnotesize
\begin{tabular}{@{}lrrrc@{}}
\toprule
Sample & Real claims & Certified & Traceable & Precision [exact 95\% CI]\\
\midrule
Development (all) & 151 & 56 & 51 & 0.91 [0.80, 0.97]\\
Confirmation (held-out) & 60 & 29 & 27 & 0.93 [0.77, 0.99]\\
\midrule
Pooled & 211 & 85 & 78 & 0.92 [0.84, 0.97]\\
\bottomrule
\end{tabular}
\end{table}

\paragraph{Which components matter} Table~\ref{tab:ablation} adds the checker's components one at a time, pooling all 211 labeled real claims (legitimate for these levels, all of which were defined before labeling). Ignoring bulk data barely changes the human-labeled results but removes a fifth of decoy matches. The context score is the component that separates traceable from misattributed claims: on its own it raises precision from 0.77 to 0.92 while keeping 60\% of traceable claims. The decoy test plays a different role. Against human labels it adds no precision on top of the context score, and it costs recall. Against values that are wrong by construction, however, it reduces false certification from 18.9\% to 1.3\%. The two components therefore guard against two different errors. The context score guards against \emph{misattribution}, a real value attached to the wrong result, which dominates the labeled samples. The decoy test guards against \emph{coincidence}, a wrong value that happens to sit near a plausible key, which is rare among labeled claims but is the failure that would matter most if an agent did fabricate a number. Users who care only about misattribution can use the score alone and gain recall; we keep both as the default because a certifier should not certify values that do not exist.

\begin{table}[t]
\centering
\caption{Component ablation on all 211 labeled real claims (158 traceable), pooling development and confirmation samples. Precision with exact 95\% intervals on unweighted counts; recall unweighted (population-weighted in parentheses). Decoy acceptance over the 1,351 development claims with at least two significant digits.}
\label{tab:ablation}
\footnotesize
\setlength{\tabcolsep}{3pt}
\resizebox{\linewidth}{!}{%
\begin{tabular}{@{}lrrcccc@{}}
\toprule
Rule & Certified & Traceable & Precision [95\% CI] & Recall & Decoy acceptance\\
\midrule
Value appears anywhere (Level 1) & 208 & 158 & 0.76 [0.70, 0.82] & 1.00 (1.00) & 87.4\%\\
\quad + ignore bulk data & 206 & 158 & 0.77 [0.70, 0.82] & 1.00 (1.00) & 71.7\%\\
\quad + context score $S(c)\ge\tau$ (no decoys) & 103 & 95 & 0.92 [0.85, 0.97] & 0.60 (0.54) & 18.9\%\\
Decoy test only, $p(c)\le\rho$ (no threshold) & 120 & 104 & 0.87 [0.79, 0.92] & 0.66 (0.51) & 3.8\%\\
\textbf{Full strict: score + decoy test} & \textbf{85} & \textbf{78} & \textbf{0.92 [0.84, 0.97]} & \textbf{0.49 (0.35)} & \textbf{1.3\%}\\
\bottomrule
\end{tabular}}
\end{table}

\paragraph{Comparison with baselines} Table~\ref{tab:baselines} compares the methods on the 129 labeled claims not used for tuning. Replacing the token-overlap score with sentence-embedding similarity makes the checker worse: tuned to the same precision, the embedding score certifies 21 claims instead of 57 (recall 0.21 against 0.54), and with the decoy test it loses precision (0.74). General-purpose sentence embeddings are a poor fit for evidence keys such as \texttt{rel\_l2\_std} or \texttt{heldout\_params\_P}, whose meaning lies in exact tokens rather than in paraphrase. Exact token overlap is what distinguishes a standard deviation from a mean, or one boundary condition from another.

The LLM verifier behaves very differently from both. Given the same evidence the annotators saw, it labels almost every traceable claim correctly (recall 0.97 on the held-out claims, 0.95 on all 211) at a precision similar to the checker's (0.89). Its weakness is the failure this paper is about: of the 53 labeled claims that are not traceable, it accepts 19 (36\%), and 19 of the 47 wrong-source claims are labeled traceable. The strict checker accepts 7 of the 53 (13\%). The LLM's agreement with the human labels is moderate ($\kappa=0.55$ over four labels, $0.64$ for traceable versus not), well below the agreement between the two annotators ($\kappa=0.89$ and $0.96$). The two methods are complementary rather than competing. When both accept a claim, 51 of 52 held-out claims are traceable (precision 0.98), and among claims the checker leaves uncertified, the LLM's ``traceable'' answers are correct 80\% of the time. A practical pipeline is therefore to let the deterministic checker certify what it can, and to use an LLM to triage the remainder for human review rather than to issue final verdicts. The checker also has properties an LLM verifier lacks: it is deterministic, costs nothing to rerun, and exposes the entry and score behind every certification. The LLM, by contrast, depends on the candidate entries that the checker's retrieval supplies.

\begin{table}[t]
\centering
\caption{Comparison with baselines on the 129 labeled real claims not used for tuning (development test split and confirmation sample; 94 traceable). Precision with exact 95\% intervals on unweighted counts. Decoy acceptance on the decoys of the 211 labeled claims; not defined for the LLM verifier, which sees only the top-ranked entries.}
\label{tab:baselines}
\footnotesize
\resizebox{\linewidth}{!}{%
\begin{tabular}{@{}lrrccc@{}}
\toprule
Method & Certified & Traceable & Precision [95\% CI] & Recall & Decoy acceptance\\
\midrule
Value appears anywhere (Level 1) & 127 & 94 & 0.74 [0.65, 0.81] & 1.00 & --\\
Embedding score (no decoys) & 21 & 20 & 0.95 [0.76, 1.00] & 0.21 & 4.7\%\\
Embedding score + decoy test & 54 & 40 & 0.74 [0.60, 0.85] & 0.43 & 3.3\%\\
LLM verifier (Claude Opus 5.5) & 102 & 91 & 0.89 [0.82, 0.94] & 0.97 & --\\
Checker \emph{and} LLM agree & 52 & 51 & 0.98 [0.90, 1.00] & 0.54 & --\\
\textbf{Checker (strict)} & \textbf{57} & \textbf{51} & \textbf{0.89 [0.78, 0.96]} & \textbf{0.54} & \textbf{1.4\%}\\
\bottomrule
\end{tabular}}
\end{table}

\paragraph{Non-verification is not evidence of falsehood} The price of this precision is recall: the strict checker certifies about a third of traceable claims (0.34 held-out, 0.39 development), and its coverage---the share of all real claims it certifies---is 23\%. The remaining claims cannot be read as failures. Among held-out claims the checker did not certify, 19 of 31 were labeled traceable (57\% population-weighted); on development the figure is 61 of 95 (72\%). A reader who treated ``not certified'' as ``not supported'' would be wrong in most cases. The relaxed operating points recover more traceable claims (recall 0.51--0.55 held-out) at a cost of about seven points of precision and a three- to seven-fold higher decoy acceptance, so we retain strict as the default and report the others as the available trade-off.

\paragraph{Why traceable claims are missed} On the development tuning split, the 37 traceable claims the strict checker missed fall into three groups. For 13, the correct entry was found and beat every decoy, but the context score fell just below the threshold, typically because a sentence reporting several numbers dilutes the overlap. The other 24 failed the decoy test; we assigned them to two groups by comparing the best-matching entry with the source the annotator identified. For 6, the context matched, but neighboring decoy values matched equally well, as happens in dense grids of similar results. For the remaining 18, a different entry scored best, most often a log line or an entry with a generic key. Two alternative scoring functions---a plain set cosine and the share of evidence-key tokens found in the claim's context---did not improve on the original (Table~\ref{tab:misses} in \ref{app:tables}). 

Recall does not depend on the agent: on human-labeled traceable claims, the strict checker certifies 50\% for sol-5.6 ($n=34$) and 49\% for all other agents combined ($n=124$; Fisher's exact $p=1.0$).

\subsection{Traceability across the benchmark}\label{sec:res-bench}

We applied the strict checker to every submission. It completed for 102: 99 contained claims and 3 contained none, while 15 exceeded a 20--40 minute time limit because of very large result files, 2 failed, and 16 exceeded the 10 GB storage limit of our environment. The 99 checked submissions contain 14,577 claims with at least two significant digits, with at least four submissions for every agent.

\begin{figure}[t]
\centering
\includegraphics[width=0.8\linewidth]{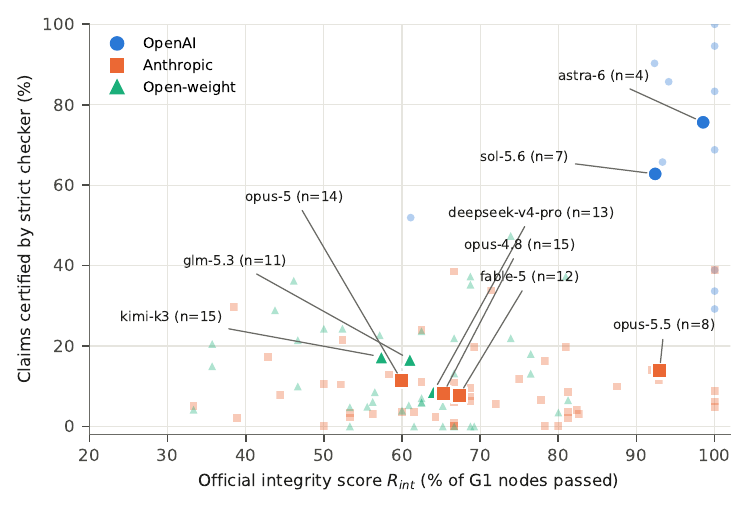}
\caption{Share of claims certified by the strict checker against the official integrity score. Small markers are submissions ($n=99$); large markers are agent means, with the number of checked submissions in parentheses. Certified shares are lower bounds on traceability, since the checker certifies about a third of traceable claims.}
\label{fig:bench}
\end{figure}

Fig.~\ref{fig:bench} plots, for each submission, the share of its claims the checker certifies against the official integrity score. The two OpenAI agents stand apart: on average the checker certifies 76\% of astra-6's claims and 63\% of sol-5.6's, against 8--17\% for each of the other seven agents. Because recall does not differ by agent (Section~\ref{sec:res-checker}), this gap reflects how traceable the agents' reported numbers are, not how easy their files are to read. Across submissions, the certified share is positively associated with the official integrity score (Spearman $\rho=0.22$, $p=0.028$, $n=99$) and negatively associated with the judge's failure rate on number-traceability nodes ($\rho=-0.32$, $p=0.002$, $n=92$). The checker and the judge use entirely different evidence---one compares numbers with files, the other reads the submission---and the association between them is positive but weak; across the nine agents it is not significant (Spearman $\rho=0.22$, $p=0.58$). Both, however, place the two OpenAI agents among the three most traceable.

Table~\ref{tab:bench} gives the per-agent figures.

\begin{table}[t]
\centering
\caption{Benchmark-wide traceability by agent (strict checker, 99 submissions). Certified: mean share of a submission's claims the checker certifies (a lower bound on traceability). Untraced: mean share of claims whose value occurs in no result file. Judge NT fail: official judge's failure rate on number-traceability nodes (all 15 submissions per agent).}
\label{tab:bench}
\footnotesize
\begin{tabular}{@{}lrrrrrr@{}}
\toprule
Agent & Subm. & Claims & Certified & Untraced & $R_\text{int}$ & Judge NT fail\\
\midrule
astra-6 & 4 & 113 & 75.6\% & 0.8\% & 98.5 & 0.0\%\\
sol-5.6 & 7 & 729 & 62.8\% & 0.8\% & 92.4 & 8.3\%\\
kimi-k3 & 15 & 1,950 & 17.3\% & 6.1\% & 57.4 & 58.3\%\\
glm-5.3 & 11 & 1,746 & 16.7\% & 5.1\% & 61.0 & 38.9\%\\
opus-5.5 & 8 & 1,206 & 13.8\% & 1.0\% & 92.9 & 13.9\%\\
opus-5 & 14 & 3,301 & 11.5\% & 4.5\% & 59.9 & 58.3\%\\
deepseek-v4-pro & 13 & 1,820 & 8.6\% & 6.6\% & 64.0 & 52.8\%\\
opus-4.8 & 15 & 1,563 & 8.2\% & 4.3\% & 65.3 & 52.8\%\\
fable-5 & 12 & 2,149 & 7.8\% & 3.2\% & 67.3 & 55.6\%\\
\midrule
All & 99 & 14,577 & & & & \\
\bottomrule
\end{tabular}
\end{table}

One agent departs from this pattern. The official judge rates opus-5.5 almost as highly as sol-5.6 (mean integrity score 93 across its checked submissions), yet the checker certifies only 14\% of its claims, a rate similar to the low-integrity agents. The difference is not one of format: 99\% of opus-5.5's values occur somewhere in its result files, and its certification rates for numbers in prose and in tables match those of the other non-OpenAI agents. The judge, however, has about 2.4 number-traceability criteria per submission, so it examines a small number of headline values, and on those opus-5.5 fails only 14\% of the time (compared with 8\% for sol-5.6 and 39--58\% for the agents without G1 passes). A manual reading of 14 uncertified opus-5.5 claims found both values that are traceable but sit in sentences dense with numbers, and values computed while writing that appear in no file. These observations are consistent with a judge that verifies headline numbers which opus-5.5 reports correctly, while much of its other numerical content is no more traceable than that of the weaker agents. An agent-specific labeled sample would be needed to confirm this, and we treat it as a hypothesis.

\subsection{Robustness to the judge (RQ4)}\label{sec:res-judge2}

The second judge completed all 30 sampled submissions. On the submissions graded more than once, each judge is highly consistent with itself (98--100\% node agreement across repeats for the second judge, 93--99\% for the official judge), so disagreements between the two reflect different judgments rather than noise. Across judges, per-repeat agreement is 82.3\% of counted nodes ($\kappa=0.51$) and 81.2\% of G1 nodes ($\kappa=0.38$), and the majority G1 verdicts agree for 22 of the 30 submissions. The disagreements are asymmetric: the second judge passes 12 submissions on G1 against the official judge's 6, passing 7 that the official judge failed and failing 1 that it passed.

\begin{table}[t]
\centering
\caption{G1 node failure rates by judge and agent vendor group on the 30 re-graded submissions (means over submissions). $d$: second judge minus official judge.}
\label{tab:vendor}
\footnotesize
\begin{tabular}{@{}lrrrr@{}}
\toprule
Agent vendor group & Submissions & Official judge & Second judge & $d$ (points)\\
\midrule
OpenAI & 8 & 3.5\% & 1.3\% & $-2.2$\\
Anthropic & 13 & 16.7\% & 8.1\% & $-8.6$\\
Open-weight & 9 & 27.6\% & 23.9\% & $-3.8$\\
\midrule
All & 30 & & & $-5.4$\\
\bottomrule
\end{tabular}
\end{table}

\begin{figure}[t]
\centering
\includegraphics[width=0.78\linewidth]{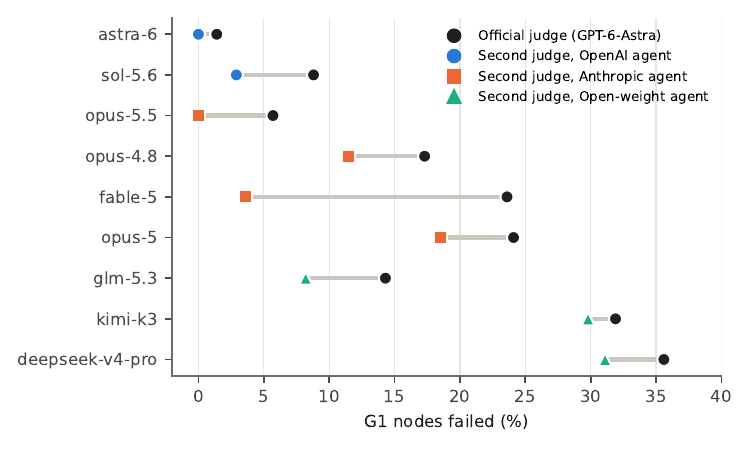}
\caption{G1 node failure rates under the official judge (black) and the second judge (colored by the agent's vendor group) on the 30 re-graded submissions, by agent. The second judge is more lenient for every agent, but both judges order the agents similarly.}
\label{fig:judges}
\end{figure}

Fig.~\ref{fig:judges} shows the comparison for each agent, and Table~\ref{tab:vendor} aggregates it by the vendor of the agent being judged.

 Two results matter for interpreting the benchmark. First, the two judge configurations order the three vendor groups identically, with the OpenAI agents' submissions cleanest by a wide margin. On this sample, then, the ordering is not specific to the official judge; but with 30 submissions and judges that differ in harness as well as model, this does not establish ranking stability across the benchmark or rule out vendor effects. Second, the second judge is more lenient overall, failing 5.4 points fewer G1 nodes per submission on average (Wilcoxon $p=0.014$). The leniency is largest for the Anthropic agents, a pattern consistent with a judge favoring its own vendor's agents, but the difference between vendor groups is not statistically significant (Kruskal--Wallis $p=0.26$; permutation test, Anthropic versus others, $p=0.16$; GEE interaction, $p=0.075$). Absolute pass rates thus depend on the judge configuration, while the vendor-group ordering was the same under both.

\section{Discussion}\label{sec:discussion}

\subsection{What the findings mean}

The four analyses point in the same direction. The official judge's failures concentrate on criteria that compare the write-up with the work, not on criteria for overt misconduct. When individual numbers are checked by hand, most unsupported ones exist in the agent's own files but belong to a different experiment, metric, or run. A provenance checker with no access to the judge's verdicts is positively, if weakly, associated with the judge's integrity scores, and both single out the two OpenAI agents. And a second judge configuration orders the vendor groups the same way, while disagreeing about how many submissions pass. In this benchmark, the typical integrity failure of an AI research agent is not invention but misattribution: the agent did the work, recorded results, and then reported some of them against the wrong referent.

This distinction matters because the two failure modes call for different responses. Fabrication is a question of intent and calls for detection and sanction. Misattribution is a question of record-keeping and reporting discipline, and calls for tooling: structured result files with explicit keys, reports generated from those files rather than retyped, and checks that tie each reported number to the run that produced it. The behavior we observe resembles the selective and loosely documented reporting that InnovationEval describes \citep{owen2026innovationeval} more than the invented results that headlines about ``fabricating'' agents suggest. It also echoes long-standing findings about human research reporting, where errors in reported statistics are common and mostly inadvertent \citep{bakker2011errors,nuijten2016statcheck}.

The checker results add a methodological point that holds regardless of how the fabrication--misattribution split varies across settings. If a reported number may be a real value attached to the wrong source, then confirming that the value appears in the outputs cannot establish that the claim is supported. Value matching accepted 87--95\% of unsupported claims in our samples and 88\% of decoy values that were wrong by construction. Any audit of AI-produced results---whether by a benchmark, a reviewer, or an organization consuming agent reports---needs to establish provenance, not presence.

\subsection{Implications for practice}

\paragraph{For organizations that use AI-produced analyses} The first implication is to ask which run produced a reported number, not whether the number can be found. The second is to use automated verification asymmetrically. A conservative checker can certify a meaningful share of claims (23\% of claims in our held-out sample, at over 90\% precision), and those claims can be relied on to the extent that the underlying experiment is sound. Claims it cannot certify should be routed to human review rather than treated as failures, since most of them turned out to be traceable. This asymmetric use guards against both failure modes of trust in automation identified in the information-systems literature: over-reliance on outputs that look verified \citep{parasuraman2010complacency,lee2004trust}, and wholesale rejection of useful work after visible errors \citep{dietvorst2015aversion}. In data-quality terms \citep{wang1996dataquality}, traceability is a dimension of fitness for use that can be measured claim by claim, and it is distinct from accuracy.

\paragraph{For benchmark designers} Integrity verdicts from a single LLM judge are stable across repeats but not across judge configurations: on our sample, the second judge passed twice as many submissions, although it ordered the vendor groups the same way. Benchmarks should therefore report which judge produced their pass rates, treat absolute pass rates as judge-dependent, and, where feasible, add a second judge from a different vendor or a deterministic check. The opus-5.5 case suggests a further risk. A judge that checks a few headline numbers per paper may rate highly an agent whose headline numbers are right but whose other numbers are not traceable. Claim-level checking complements node-level judging rather than replacing it.

\paragraph{For agent developers} Our failure analysis suggests that traceability depends on how results are recorded: entries with descriptive keys are easy to tie to the sentences that report them, while values computed only in the prose cannot be traced at all. Making reports a function of the recorded results---for example, by inserting values from result files instead of retyping them---would remove most of the misattribution we observed, and would make agents' work far cheaper to verify.

\subsection{Limitations}

\paragraph{One benchmark} All findings come from one benchmark, nine agents, and fifteen tasks in the computational and physical sciences. The share of misattributed versus absent numbers may differ in other domains, with other agents, or under other harnesses, and we make no claim that agents in general misreport more often than they fabricate. The checker relies on properties of this benchmark---papers written in \LaTeX{} and result files in machine-readable formats---and would need adaptation for papers in other formats or results stored in binary files or databases.

\paragraph{Labels and annotators} The human labels come from two annotators, both members of the research team. They worked blind to the checker and to each other's labels, and their agreement is high, but they are not an external panel. The development labeling rules were refined while labeling, before the confirmation sample was drawn. During confirmation labeling, four labels were revised after a rule-consistency review, and we report agreement before and after. ``Not found'' means not found in the available artifacts: a value computed and never saved, or quoted from the literature, is not fabricated. The labeled samples are modest (256 claims, 60 real claims in the held-out set), and several confidence intervals are correspondingly wide.

\paragraph{Checker scope} The checker certifies provenance, not correctness. A certified number is supported by a recorded result, but the experiment that produced it may still be flawed. Its recall is about one third, so it settles only part of a paper automatically. Benchmark-wide measurements cover 99 of the 135 submissions; the excluded submissions are the largest, and if very large runs differ systematically in how they report results, the per-agent certified shares could shift.

\paragraph{Second judge and harnesses} The second-judge sample has 30 submissions, most graded once, so the vendor-interaction tests have limited power, and the non-significant hint of self-preference can be neither confirmed nor ruled out. The two judge configurations also differ in execution harness and file-access restrictions, not only in model, so even a significant difference could not be attributed to the judge's vendor alone. We also did not adjudicate the 438 nodes on which the two judges disagree, so we can say that pass rates depend on the judge but not which judge is closer to a human expert. Agents were run under different harnesses. For example, the runner's completion check failed to execute much more often for some Claude-based agents than for others, so differences between agents may partly reflect their harnesses as well as their models.

\paragraph{The agents themselves} The agents in the benchmark are recent frontier systems that will be superseded. We expect the methodological findings---that presence is not provenance, and that non-verification is not falsehood---to outlast the specific rates we report.

\section{Conclusions}\label{sec:conclusion}

AI research agents now produce papers that look finished, and the people who read them cannot easily see how each reported number was obtained. We examined the integrity failures of nine such agents on an open benchmark using the official judgments, claim-level human labels, a deterministic claim-to-evidence checker, and a second judge. In this benchmark, unsupported numbers are mostly real values reported against the wrong experiment, metric, or run, rather than values with no source. Because such numbers exist in the agents' own files, checking whether a value appears accepts almost all of them. A checker that instead establishes provenance certifies claims with high precision on held-out data. But most claims it cannot certify are traceable, so failure to verify must not be read as evidence of falsehood. Applied across the benchmark, the checker is positively associated with the official judge's integrity scores, and on a sample a second judge configuration orders the agents' vendor groups the same way while passing twice as many submissions, so pass rates depend on the judge. The checker's limited recall restricts automation to a subset of claims; it does not limit the value of verifying that subset. Verifying AI-produced research therefore means tracing each number to the result it claims to describe. The tools, labels, and data we release make that possible for a substantial share of claims today.

\appendix
\section{Supplementary tables}\label{app:tables}
\setcounter{table}{0}

\begin{table}[t]
\centering
\caption{Integrity score and run length: OLS estimates of Eq.~\eqref{eq:fe} with HC1 standard errors ($n=135$ submissions).}
\label{tab:fe}
\footnotesize
\begin{tabular}{@{}lrrr@{}}
\toprule
& (1) No fixed effects & (2) Agent FE & (3) Agent + task FE\\
\midrule
$\log h$ (hours) & $-5.21$ & $-0.33$ & $-0.61$\\
& (1.28) & (0.99) & (1.12)\\
$p$-value & $<0.001$ & 0.74 & 0.59\\
Agent fixed effects & -- & \checkmark & \checkmark\\
Task fixed effects & -- & -- & \checkmark\\
$R^2$ & 0.11 & 0.61 & 0.77\\
\bottomrule
\end{tabular}
\end{table}

\begin{table}[t]
\centering
\caption{Why the strict checker misses traceable claims: the 37 traceable claims it did not certify on the development tuning split.}
\label{tab:misses}
\footnotesize
\begin{tabular}{@{}>{\raggedright\arraybackslash}p{4.6cm}r>{\raggedright\arraybackslash}p{6.2cm}@{}}
\toprule
Cause & Claims & What happens\\
\midrule
Score just below threshold & 13 & Correct entry found and beats every decoy; context overlap diluted by sentences reporting several numbers\\
Decoys match equally well & 6 & Dense grids of similar results with similar keys; $p(c)>0.1$\\
Another entry scores best & 18 & A log line, a generic key, or a near-duplicate entry outranks the correct one\\
\midrule
Total & 37 & \\
\bottomrule
\end{tabular}
\end{table}

\section*{Data and code availability}
All code, human labels, derived data, and figures are publicly available\ifanon\ (repository link withheld for double-blind review)\else\ at \url{https://github.com/mos3b-faqeeh/agent-integrity-analysis}\fi. The benchmark data are available from the original authors \citep{kim2026unknown}; we do not redistribute raw submissions.

\section*{Declaration of generative AI and AI-assisted technologies in the writing process}
During the preparation of this work the authors used Claude (Anthropic) to assist with data analysis code, drafting and editing the manuscript, and a consistency review of annotation labels against the written labeling rules. After using this tool, the authors reviewed and edited the content as needed and take full responsibility for the content of the publication.

\section*{Declaration of competing interest}
The authors declare that they have no known competing financial interests or personal relationships that could have appeared to influence the work reported in this paper.

\bibliographystyle{elsarticle-harv}
\bibliography{refs}
\end{document}